\documentclass[10pt,twocolumn,letterpaper]{article}

\usepackage{cvpr}              

\usepackage[dvipsnames]{xcolor}
\usepackage{adjustbox}
\usepackage{cuted}
\usepackage{pifont}
\newcommand{\red}[1]{\textbf{\textcolor{red}{#1}}}
\newcommand{\blue}[1]{\underline{\textcolor{blue}{#1}}}

\renewcommand{\paragraph}[1]{\vspace{-0mm}\noindent\textbf{#1}\hspace{0mm}}
\newcommand{\Subsection}[1]{\subsection{#1} \vspace{-0mm}}

\definecolor{cvprblue}{rgb}{0.21,0.49,0.74}
\usepackage[pagebackref,breaklinks,colorlinks,citecolor=cvprblue]{hyperref}

\def\paperID{9785} 
\def\confName{CVPR}
\def\confYear{2024}

\title{Perception-Oriented Video Frame Interpolation via Asymmetric Blending}

\author{
Guangyang Wu$^{1}$\quad
Xin Tao$^{2}$\quad 
Changlin Li$^{3}$\quad
Wenyi Wang$^{4}$\quad
Xiaohong Liu$^{1 \dagger}$\quad
\stepcounter{footnote}Qingqing Zheng$^{5}$\thanks{~Corresponding authors.}\\
$^1$Shanghai Jiao Tong University\quad
$^2$Kuaishou Technology\quad
$^3$SeeKoo\\
$^4$University of Electronic Science and Technology of China\\
$^5$Shenzhen Institute of Advanced Technology, Chinese Academy of Sciences
}

\begin{document}
\maketitle
\begin{strip}
\begin{minipage}{\textwidth}\centering
\vspace{-30pt}
\begin{picture}(0,0)
    \put(-200,-110){\makebox(0,0)[bc]{\small Ground-Truth}}
    \put(-100,-110){\makebox(0,0)[bc]{\small RIFE~\cite{huang2020rife}}}
    \put(0,-110){\makebox(0,0)[bc]{\small EMA-VFI~\cite{zhang2023extracting}}}
    \put(100,-110){\makebox(0,0)[bc]{\small STMFNet~\cite{Danier_2022_CVPR}}}
    \put(200,-110){\makebox(0,0)[bc]{\small PerVFI (ours)}}
\end{picture}
\includegraphics[width=\linewidth, height=0.2\linewidth]{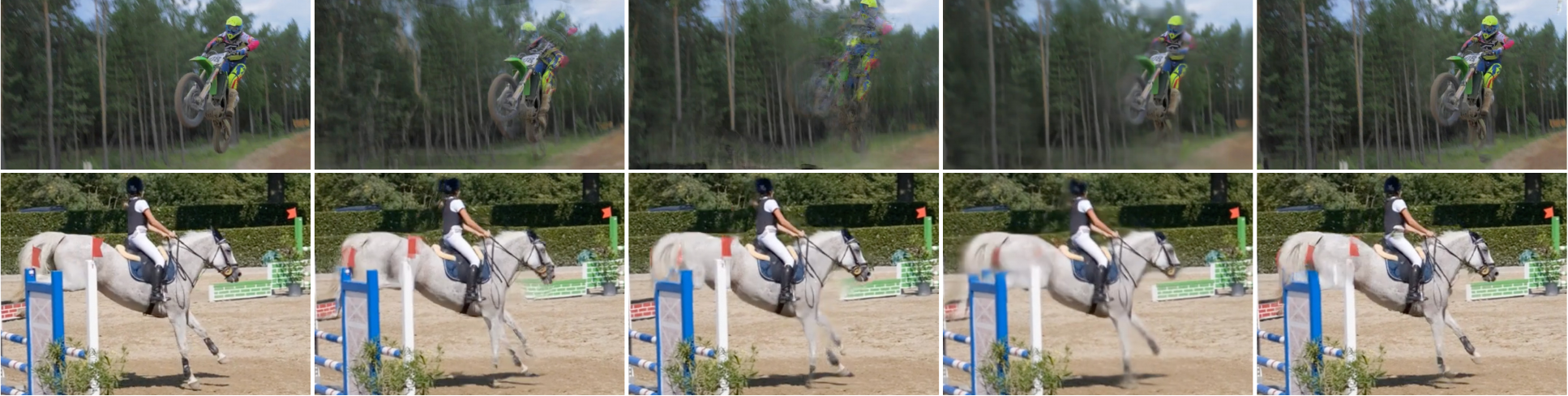}
\vspace{-5pt}
\captionof{figure}{We present challenging video frame interpolation examples, demonstrating our approach excels in handling large motion, outperforming alternatives prone to blurriness or ghosting.}\label{fig:teaser}
\end{minipage}
\end{strip}


\begin{abstract}
        Previous methods for Video Frame Interpolation (VFI) have encountered challenges, notably the manifestation of blur and ghosting effects. These issues can be traced back to two pivotal factors: unavoidable motion errors and misalignment in supervision. In practice, motion estimates often prove to be error-prone, resulting in misaligned features. Furthermore, the reconstruction loss tends to bring blurry results, particularly in misaligned regions. To mitigate these challenges, we propose a new paradigm called PerVFI (Perception-oriented Video Frame Interpolation). Our approach incorporates an Asymmetric Synergistic Blending module (ASB) that utilizes features from both sides to synergistically blend intermediate features. One reference frame emphasizes primary content, while the other contributes complementary information. 
        To impose a stringent constraint on the blending process, we introduce a self-learned sparse quasi-binary mask which effectively mitigates ghosting and blur artifacts in the output. 
        Additionally, we employ a normalizing flow-based generator and utilize the negative log-likelihood loss to learn the conditional distribution of the output, which further facilitates the generation of clear and fine details. Experimental results validate the superiority of PerVFI, demonstrating significant improvements in perceptual quality compared to existing methods. Codes are available at \url{https://github.com/mulns/PerVFI}
 \vspace{-20pt}
 \end{abstract}

\section{Introduction}
Video frame interpolation (VFI) is an important task in computer vision that focuses on synthesizing intermediate frames between consecutive frames in a video sequence. This technique plays a crucial role in various applications such as video enhancement~\cite{toflow,fastllve}, slow-motion rendering~\cite{zoomingslomo, superslomo2018, accflow}, and frame rate conversion~\cite{hevc}, especially for producing high-definition videos~\cite{huang2020rife, xvfi2021}.

Recently, the VFI task has also derived benefits from the application of deep neural networks~\cite{superslomo2018, qvi2019, eqvi2020, softmaxsplat2020, contextaware2018, xvfi2021, huang2020rife,RPSRMD,pred,Jin_2023_WACV, toflow, bmbc, ifrnet, vsr,adaptive-btv,vsr-lut, hvfi, raw-vsr, stsr, vfiformer, vfigdc}. 
We broadly posit that recent methods typically consist of 3 main modules. The motion estimation module is used to estimate motion between consecutive frames using optical flow or deformable kernels. Subsequently, the alignment and fusion module aligns reference frames via warping operator or deformable convolution. Finally, the reconstruction module generates final results from extracted features. Despite the success of recent methods, blurry results and ghosting artifacts persist as inevitable problems. (shown in Figure~\ref{fig:teaser}). We attribute this primarily to two inherent challenges:


\paragraph{Inevitable Motion Errors.}
Ideally, with accurate motion estimates, the aforementioned procedure can yield satisfactory results. However, achieving error-free pixel-wise correspondence for real-world videos proves challenging, especially in the presence of large-scale motions. Unlike several prior methods~\cite{Jin_2023_WACV, abme2021, xvfi2021, Danier_2022_CVPR} that primarily focus on enhancing the quality of motion estimation, our objective is to fortify our network against alignment errors. Specifically, after a thorough investigation of existing methods, we conclude that in cases of inaccurate motion estimation, the network struggles to discern the correct frame. Consequently, preceding algorithms often produce blurred and ghosted results by averaging multiple frames. We contend that basing the output on a single frame while utilizing other frames to supplement specific details holds the potential to yield clearer and more plausible results.

\paragraph{Temporal Supervision Misalignment.} 
Another crucial yet often overlooked issue in VFI is the temporal uncertainty. During the training phase, the ground truth (GT) intermediate frame only provides a reference at a specific time. However, in the case of a continuous natural video, scenes captured in the time interval between two frames can offer multiple potential solutions. Therefore, the learned intermediate features can vary across different training videos. We term this issue Temporal Supervision Misalignment, and this misalignment may cause the network to produce blurry results. To address this problem, conventional pixel-wise loss functions such as L1 and L2 are inadequate. Instead, we opt for generative models to reconstruct results sampled from a distribution.

In order to tackle the aforementioned challenges, we propose a novel perception-oriented video frame interpolation paradigm in this paper, referred to as PerVFI. Our approach introduces an Asymmetric Synergistic Blending (ASB) module and a self-learned sparse quasi-binary mask to fuse multi-frame features. In this process, one reference frame emphasizes primary content, while the other provides complementary information. Additionally, we employ a normalizing flow-based generator to decode the intermediate features. This generator models the conditional distribution of the output based on the reference inputs. Unlike GAN-based methods that struggle to converge~\cite{Danier_2022_CVPR} and diffusion-based methods~\cite{danier2023ldmvfi} with numerous iterations, our normalizing flow-based approach demonstrates stability during training and low latency during inference. The proposed PerVFI paradigm consistently produces visually high-quality results, even in the presence of misalignment due to inaccurate motion estimates.

The contributions of this paper are as follows:
\begin{itemize}[]
    \item We introduce a novel paradigm called PerVFI, specifically designed for the perception-oriented task of VFI. Our proposed method tackles the issue of misalignment by incorporating an asymmetric synergistic blending module (ASB) and a conditional normalizing flow-based generator.
    \item To effectively control the blending process in ASB, we propose a novel quasi-binary mask. This mask allows for sparse confidence values overall and adaptive values for partial areas, effectively addressing the occlusion and imposing a strict constraint on blending process.
    \item We have conducted extensive experiments to validate the efficacy of the proposed PerVFI. The experimental results demonstrate that the PerVFI paradigm is capable of generating visually plausible outputs even in the presence of inaccurate motion estimates, exhibiting state-of-the-art performance in terms of perceptual quality.
\end{itemize}

\section{Related Works}
\label{sec:related}

\Subsection{Video frame interpolation.}
Existing VFI approaches are mostly based on deep learning~, and can be generally categorized as optical flow-based or kernel-based. Optical flow-based methods rely on optical flow estimation to generate interpolated frames\cite{superslomo2018, qvi2019, eqvi2020, softmaxsplat2020, contextaware2018, xvfi2021, huang2020rife, Jin_2023_WACV, toflow, bmbc, abme2021, ifrnet}. On the other hand, kernel-based methods argue that optical flows can be unreliable in dynamic texture scenes~\cite{adacof2020, edsc2020, cdfi2021, sepconv2017, dsepconv2020, featureflow, vfiformer}, so they predict locally adaptive convolution kernels to synthesize output pixels. Other than these two classes, there are also attempts to combine flows and kernels~\cite{dain2019, memc2021, Danier_2022_CVPR,hvfi} and to perform end-to-end frame synthesis~\cite{cain, flavr}.
It is noted that the above methods use symmetric blending, which easily blends the features from two sides with equal contribution, even when they are misaligned. Although these results in reasonably good PSNR performance, it has been previously reported~\cite{danier22icip} that PSNR does not fully reflect the perceptual quality of interpolated videos, exhibiting poor correlation performance with subjective ground truth. To improve perceptual performance, some existing methods~\cite{softmaxsplat2020, contextaware2018} use the perceptual loss~\cite{perceptual} in combination with the L1 loss. An alternative approach uses GANs~\cite{Danier_2022_CVPR, adacof2020} to enhance perceptual quality of interpolated videos. However, due to the instability of GAN training, these models are pretrained using L1 loss before fine-tuned with adversarial loss, leading to limited improvement in perceptual quality.

\Subsection{Normalizing Flow-based model.}
Recently, normalizing flow-based models have shown remarkable performance in synthesizing high-fidelity images and videos~\cite{srflow, glow, llieflow, realnvp}. These models leverage an invertible network to establish a mapping from a complex distribution to a simple distribution. Users can then decode a latent code sampled from the simple distribution to the target domain. Normalizing flow-based methods have been reported to outperform GANs in image generation tasks~\cite{llieflow, srflow, glow}. To the best of our knowledge, we are the first to adopt a normalizing flow-based model for VFI. In particular, we utilize conditional normalizing flow-based models~\cite{srflow}, which have demonstrated a strong ability to synthesize images with conditional information.

\section{Proposed Method: PerVFI}

\begin{figure*}
    \centering
    \includegraphics[width=\linewidth]{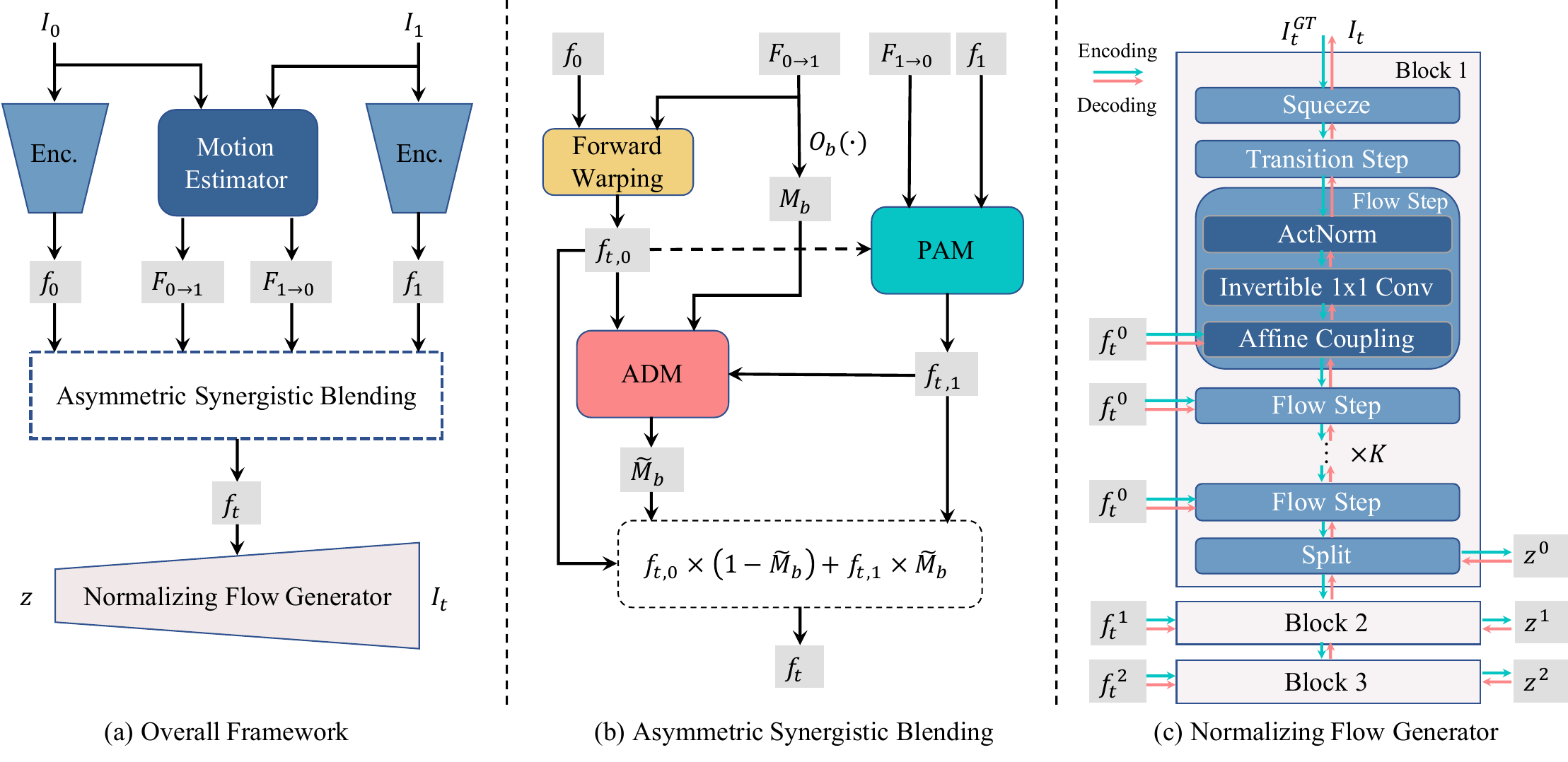}
    \caption{(a): Overview of the entire PerVFI framework. (b): Structure of the proposed Asymmetric Synergistic Blending (ASB) module. (c): Structure of the conditional normalizing flow-based generator.}
    \label{fig:overall}
    \vspace{-5pt}
\end{figure*}
Given two referenced frame images $I_0$ and $I_1 \in \mathbb{R}^{H\times W\times 3}$ with height $H$ and width $W$, our goal is to reconstruct the intermediate frame $I_t$ regarding the target time $t\in (0,1)$. The overall framework of PerVFI is presented in Figure~\ref{fig:overall}-(a), which includes an asymmetric synergistic blending (ASB) module illustrated in Figure~\ref{fig:overall}-(b) and a conditional normalizing flow-based generator illustrated in Figure~\ref{fig:overall}-(c). 

Our first step is to estimate bidirectional optical flows, denoted by $\mathbf{F}_{0 \rightarrow 1}$ and $\mathbf{F}_{1 \rightarrow 0}$, using a motion estimator such as RAFT~\cite{raft} or GMFlow~\cite{gmflow}.
Concurrently, we use a pyramidal architecture that extracts features at different scales to capture multiscale information. Specifically, we encode the two images into pyramid features with $L$ levels using a feature encoder $\mathcal{E}_\theta$, denoted as $f_i = \mathcal{E}_\theta(I_i)$ for $ i=0,1$. 
Once the bidirectional optical flow and feature pyramid have been obtained. We utilize a feature blending module, denoted as $\mathcal{B}_\theta$ and obtain intermediate pyramid features by blending, denoted as $f_t = \mathcal{B}_\theta(t, f_0, f_1, \mathbf{F}_{0 \rightarrow 1}, \mathbf{F}_{1 \rightarrow 0})$. 
Then, we decode $f_t$ into the output frame $I_t$ using a conditional normalizing flow-based generator $\mathcal{G}_\theta$ which is invertible, denoted as $I_t = \mathcal{G}^{-1}_\theta(z; f_t)$ where $z\sim \mathcal{N}(0,\tau )\in \mathbb{R}^{H\times W\times 3}$  is a variable sampled from a normal distribution with temperature $\tau$.
The feature pyramid, $f_t = \{f_t^l \text{ }|\text{ } l\in[0,1,\dots,L-1]\}$ represents $L$ features with shapes of $\frac{H}{2^l}\times \frac{W}{2^l}$.

\Subsection{Asymmetric Synergistic Blending}
\label{subsec:blending}
To effectively learn $f_t$ immune to  bidirectional motion misalignment, an intuitive insight  involves extracting primary information from one reference frame and compensating for occlusion information from the other frame, instead of simply averaging the information from both frames without any constraints. This can be achieved by obtaining a binary occlusion mask that describes whether certain regions are occluded or not, and blending the aligned features from both sides using this mask. However, due to the inaccurate motion estimates, obtaining an accurate binary occlusion mask is challenging, and aligning the features from both sides is non-trivial. Therefore, we propose a novel Asymmetric Synergistic Blending (ASB) module, which focuses on addressing these two problems.
The ASB module consists of two major components: the Pyramid Alignment Module (PAM) and the Adaptive Dilation Module (ADM). The PAM is designed to achieve more accurate alignment of features from both sides, while the ADM aims to provide a quasi-binary mask that can serve as a weighting mask for better handling of occlusion.

\paragraph{Pyramid Alignment Module.}
Alignment is a crucial step in our framework, involving the warping of reference frames and aligning pyramid features. We employ different warping operators: backward warping ($\overleftarrow{\omega}$) and forward warping with different splatting methods. For multiple-to-one situations~\cite{softmaxsplat2020}, we use the average splatting operator ($\overrightarrow{\omega_{avg}}$) and the softmax splatting operator ($\overrightarrow{\omega_{Z}}$) which subjects to an importance metric $Z$.

Following \citet{softmaxsplat2020}, the $f_0$ is warped to time $t$ using a Forward Warping module with a small neural network $v_\theta$. The importance metric $Z$ is computed as:
\begin{equation}
    Z = v_\theta(f_0^0, -\left\lVert f_0^0-\overleftarrow{\omega}(f_1^0, \mathbf{F}_{0 \rightarrow 1}) \right\rVert ),
\end{equation}
and the warped pyramid features $f_{t,0}$ are obtained through:
\begin{equation}
    f_{t,0}^l = \overrightarrow{\omega_{Z}}(f_0^l, t, \mathbf{F}_{0 \rightarrow 1}^l; Z^l), \quad l=0,1, \dots, L-1
\end{equation}
where $\mathbf{F}_{0 \rightarrow 1}^l$ and $Z^l$ are spatially downscaled optical flow and metric, respectively, by a factor of $2^l$.

To handle occlusion regions in the warped features $f_{t,0}$, we introduce a Pyramid Alignment Module (PAM) to align $f_1$ to the target time $t$, producing $f_{t,1}$. The PAM utilizes a multiscale deformable convolution network $u$ to perform alignment. Instead of warping $f_1$ directly using $\mathbf{F}_{1 \rightarrow t}$, which may cause misalignment because of inaccurate bidirectional optical flows, the PAM converts $\mathbf{F}_{1 \rightarrow t}$ to an initialized offset to guide the alignment process. This approach improves the handling of occlusion regions, resulting in more accurate alignment. The alignment process is formulated as:
\begin{align}
    &\mathbf{F}_{1 \rightarrow t} = (1-t)\cdot \mathbf{F}_{1 \rightarrow 0}, \\
    &f_{t,1} = u_\theta(-1 \times \overrightarrow{\omega_{avg}}(\mathbf{F}_{1 \rightarrow t}, \mathbf{F}_{1 \rightarrow t}),  f_1, f_{t,0}).
\end{align}
The network $u_\theta$ iteratively refines the transformation parameters in a coarse-to-fine manner, starting with aligning the features at the coarsest scale and propagating the alignment to finer scales. This helps to handle large pixel displacements and achieve more accurate alignment of features at different scales. A detailed network structure of $u_\theta$ and $v_\theta$ can be found in the appendix.

\paragraph{Adaptive Dilation Module.} 
To mitigate errors caused by binary masks generated from motion estimates and improve occlusion compensation, we introduce a lightweight Adaptive Dilation module (ADM). This module generates a quasi-binary weighting mask to control the blending of features from two sides. Specifically, ADM adaptively dilates the binary mask with convolution layers and maintains the sparsity property. In ADM, we independently generate one weighting mask for each pyramid level.
Before dilation, we obtain the binary occlusion mask $M_b^l \in \mathbb{R}^{\frac{H}{2^l}\times \frac{W}{2^l}}$ at level $l$ by applying a threshold method, denoted as $\mathcal{O}_b$, to $\mathbf{F}^l_{0 \rightarrow 1}$:
\begin{equation}\label{eq:5}
    M^l_b = \mathcal{O}_b(\mathbf{F}^l_{0 \rightarrow 1}).
\end{equation}
For each pixel $\mathbf{x}\in {(1,\dots, \frac{H}{2^l})}\times{(1,\dots,\frac{W}{2^l})}$, we compute:
\begin{equation}\label{eq:6}
    M^l_b(\mathbf{x}) = 
    \begin{cases}
        1 , &\text{if  } \overrightarrow{\omega_{avg}}(\mathit{M}_1^l, t\cdot \mathbf{F}^l_{0 \rightarrow 1})(\mathbf{x}) < \epsilon; \\
        0,  &\text{otherwise}.
    \end{cases}
\end{equation}
Here, $M_1^l$ is a constant metric with the same shape as $M_b^l$ filled with value $1$, and the threshold $\epsilon$ is a constant set to $0.5$ by default.

\begin{figure*}[t]
    \centering
    \includegraphics[width=\linewidth]{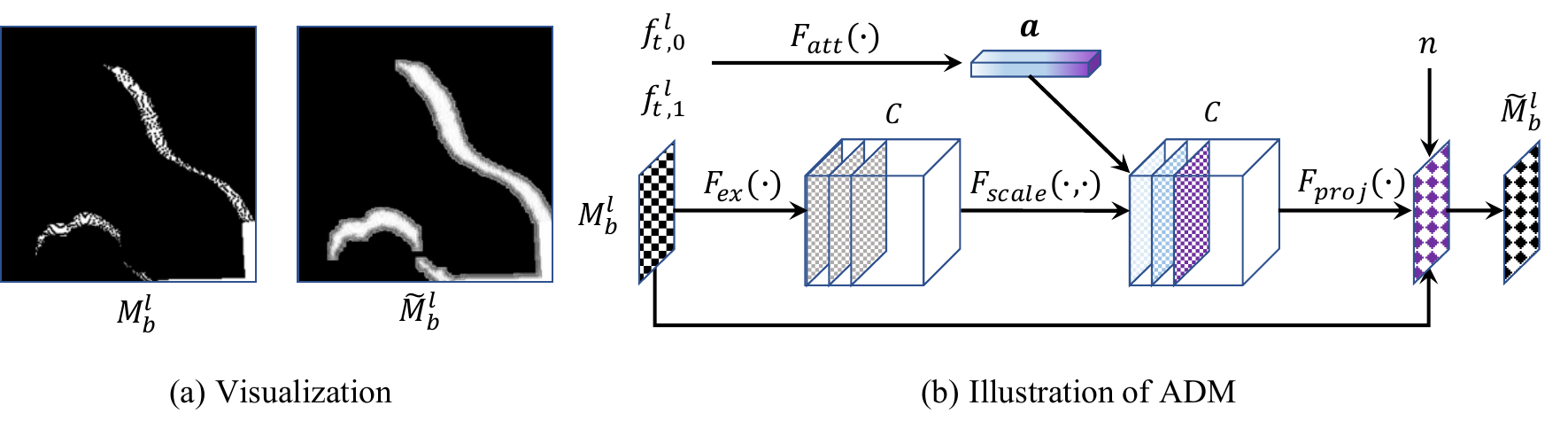}
    \caption{The Adaptive Dilation Module (ADM) produces the quasi-binary mask $\widetilde{M}^l_b$ by leveraging the binary occlusion mask $M_b^l$ and the two aligned feature sets $f_{t,0}$ and $f_{t,1}$. Panel (a) provides a visualization of the input and output masks, while panel (b) presents the flowchart outlining the operations of ADM. Further information regarding the intricacies of the module is detailed in Equations~\ref{eq:5} - \ref{eq:8}.}
    \label{fig:quasibinary}
    \vspace{-10pt}
\end{figure*}

ADM generates the quasi-binary mask $\widetilde{M}^l_b$ based on $M_b^l$ and the two aligned feature sets $f_{t,0}$ and $f_{t,1}$, as illustrated in Figure~\ref{fig:quasibinary}.
Firstly, we expand $\mathcal{M}_b$ to $C$ feature maps using convolution layers denoted as $\mathcal{F}_{ex}$, consisting of three bias-free convolutional layers with kernel sizes $(7,3,1)$, respectively.
The combination of these layers results in a $17\times 17$ dilation field for each occluded pixel in $M_b^l$.
We then obtain a sample-dynamic attention weight $\mathbf{a}\in \mathbb{R}^C$ derived from reference features using squeeze and excitation layers, denoted as $\mathcal{F}_{att}$. The expanded features are scaled with normalized attention weights, denoted as $\mathcal{F}_{scale}$. Afterwards, the scaled features are projected to metric $\widehat{M}^l$ through one bias-free convolution layer with kernel size $1$, denoted as $\mathcal{F}_{proj}$. The procedure can be formulated as:
\begin{equation}\label{eq:7}
    \widehat{M}^l = \mathcal{F}_{proj}( \mathcal{F}_{scale}( \mathcal{F}_{exp}(M_b^l),  \mathcal{F}_{att}(f_{t,0}^l, f_{t,1}^l))).
\end{equation} 
Finally, the quasi-binary mask is obtained using the following equation:
\begin{equation}\label{eq:8}
    \widetilde{M}_b^l = \tanh(\text{abs}(\widehat{M}^l + \alpha\cdot n) + \beta\cdot M_b^l).
\end{equation}
Here, $\beta$ controls the salience of occluded regions and set to $2$ by default. $n\sim \mathcal{U}(-1,1)$ is a random noise with the same shape as $M_b^l$, and $\alpha$ is set to $1e^{-3}$ during training and $0$ during inference. We found it necessary to add the random noise during training to avoid gradient vanishing and lead to more robust occlusion compensation. By blending the two sides information through quasi-binary mask, we obtain the output pyramid feature $f_t$ as follows:
\begin{equation}
    f_t^l = f^l_{t,0} \cdot (1 - \widetilde{M}_b^l) + f^l_{t,1} \cdot \widetilde{M}_b^l.
\end{equation}
The adaptive dilation field of the quasi-binary mask maintains its sparsity and reduces the impact of errors in the binary mask, therefore providing more plausible primary content. Moreover, the sample-dynamic attention mechanism in the ADM module adapts to different pyramid levels and reference features, providing more robust occlusion compensation.

\Subsection{Normalizing Flow Generator}
\label{subsec:normal}

To parameterize the conditional distribution $p(I_t|f_t)$, we utilize an invertible neural network $\mathcal{G}_{\theta}$, which maps the intermediate feature $f_t$ and the target image $I_t$ to a latent variable $z=\mathcal{G}_{\theta}(I_t; f_t)$. By employing an invertible normalizing flow-based generator $\mathcal{G}_{\theta}$, we ensure that $I_t$ can be accurately reconstructed from the latent encoding $z$ as $y=\mathcal{G}_{\theta}^{-1}(z; f_t)$. By assuming a simple distribution $p_z(z)$ (\textit{e.g.}, Gaussian) in the latent space $z$, the distribution $p(I_t| f_t, \theta)$ is implicitly defined through the mapping $y=\mathcal{G}_{\theta}^{-1}(z; f_t)$ of $z\sim p_z$.
In a normalizing flow-based generator, the probability density $p$ can be explicitly computed as:
\begin{equation}
    p(I_t|f_t, \theta) = p_z(\mathcal{G}_\theta(I_t; f_t))\left\lvert \det\frac{\partial \mathcal{G}_\theta}{\partial I_t}(I_t; f_t)\right\rvert,
\end{equation}
and we can train the parameters $\theta$ by minimizing the negative log-likelihood (NLL). Let $\hat{p} = -\log p_z(\mathcal{G}_\theta(I_t; f_t))$, the NLL is formulated as:
\begin{align}
    \mathcal{L}_{nll}(\theta; f_t, y_t) &= -\log p(I_t|f_t, \theta) \nonumber\\
    &= \hat{p} - \log \left\lvert \det \frac{\partial \mathcal{G}_\theta}{\partial I_t}(I_t; f_t) \right\rvert.
\end{align}
As shown in Figure~\ref{fig:overall}-(c), we decompose $\mathcal{G_\theta}$ into a sequence of $N$ invertible layers $h^{n+1}=\mathcal{G}_\theta^n(h^n;f_t)$, where we have $h^0=I_t$ and $h^N=z$. The invertible layers consist of Squeeze Layer, Transition Layer, Invertible 1x1Conv Layer, Actnorm Layer, Affine Coupling Layer, and Split Layer.
By applying the chain rule along with the multiplicative property of the determinant, the NLL can be expressed as:
\begin{equation}\small
    \mathcal{L}_{nll}(\theta; f_t, I_t) = \hat{p} - \sum_{n=0}^{N-1}\log \left\lvert\det \frac{\partial \mathcal{G}_\theta^n}{\partial h^n}(h^n; f_t) \right\rvert.
\end{equation}
We thus only need to compute the log-determinant of the Jacobian $\frac{\partial \mathcal{G}_\theta^n}{\partial h^n}$ for each individual flow-layer $\mathcal{G}_\theta^n$. Other than NLL loss, we also adopt perceptual loss as auxiliary loss implemented with VGG-network~\cite{johnson2016perceptual}. We find that the introduction of auxiliary loss can significantly improve the convergence speed of the network training, and the final generated images are less noisy and clearer. The auxiliary loss is calculated as follows:
\begin{equation}\small
    z'\sim \mathcal{N}(mean(\mathcal{G}_\theta(I_t^{GT}; f_t)), var(\mathcal{G}_\theta(I_t^{GT}; f_t))),
\end{equation}
\begin{equation}\small
    \mathcal{L}_{per}(\theta; f_t, z') = ||\mathcal{F}_{vgg}(\mathcal{G}_\theta^{-1}(z'; f_t)) - \mathcal{F}_{vgg}(I_t^{GT})||_2.
\end{equation}
Here, we first encode the $I_t$ to latent space $z$, then randomly sample $z'$ according to the mean and variation of $z$ for decoding. The final bidirectional loss is formulated as:
\begin{equation}
\label{eq:loss}
    \mathcal{L}(\theta; f_t, I_t) = \mathcal{L}_{nll} + \mu\cdot\mathcal{L}_{per}.
\end{equation}

\paragraph{Affine Coupling.} The affine coupling layer integrate the condition information and easily invertible. Unlike the conditional affine coupling layer introduced in~\cite{srflow}, where the inverse operation is afflicted by numerical instability, we modify it into a more stable version as follows:
\begin{gather}
    h_A^{n+1} = h_A^n, \nonumber\\
    h_B^{n+1} = \exp(\lambda^n\cdot\tanh(w_s^n(h_A^n; f_t))+ \eta^n)\cdot h_B^n \nonumber\\ + w_b^n(h_A^n; f_t). \label{eq:coupling}
\end{gather}
Here, $h^n = (h_A^n, h_B^n)$ is a partition of the feature map in the channel dimension. $w_s^n$ and $w_b^n$ are neural networks generating the scaling and bias of $h_B^n$. $\lambda^n$ and $\eta^n$ are learnable scalars. For stability, we initialize parameters of $\lambda^n$ and $\eta^n$ to $1$, and the last convolutional layer of $w_s^n$ and $w_b^n$ to $0$. Since the Jacobian of~(\ref{eq:coupling}) is triangular, its log-determinant is easily calculated as 
\begin{equation}
\lambda^n\cdot\sum_{ijk}(\tanh(w_s^n(h_A^n; f_t))_{ijk} + \sum_{ijk}\eta^n.
\end{equation}

\paragraph{Other Details.} The Invertible ${1 \times 1}$ Conv layer, ActNorm Layer, Squeeze Layer and Transition Layer are following the basic settings in~\cite{srflow}. Specifically, we stack $L=3$ blocks regarding three pyramid levels, each containing $K=16$ flow-steps. During encoding, in each block, the Split Layer outputs a latent variable $z_l$, and the final latent variable $z = (z_l)_{l=1}^L$ models variations in the image at different resolutions. During decoding, all the components in $z$ are independently sampled. More details can be found in the appendix.

\section{Experiments}
\label{sec:exp}

\Subsection{Experimental Settings}
\label{subsec:setting}
\paragraph{Training Settings.} The PerVFI network is trained using the bidirectional loss in Equation~(\ref{eq:loss}), with $\mu$ set to $0.2$ as the default value. It is worth noting that during training, we enhance the robustness of our network by randomly selecting motion estimates generated by RAFT~\cite{raft} and GMFlow~\cite{gmflow}. For motion estimation modules, we utilize the official pretrained models on the Sintel dataset and freeze their parameters. Our training dataset consists of frame-triples from the training portion of the publicly available Vimeo-90k~\cite{toflow} dataset. Throughout the training process, we employ a patch size of $256\times 256$, with a batch size of 16 for each iteration. The ADAM optimizer with default hyperparameter settings in~\cite{adam} is used, and the initial learning rate is set to $5e^{-4}$. The learning rate is halved every 20 epochs. The PerVFI model is trained until convergence, which typically occurs around 64 epochs, using two NVIDIA RTX 3090 GPUs.



\paragraph{State-of-the-art Methods.} We compare our approach to several state-of-the-art video frame interpolation methods, including EDSC~\cite{cheng2021multiple}, RIFE~\cite{huang2020rife}, VFIFormer~\cite{vfiformer}, EMA-VFI~\cite{zhang2023extracting}, AMT~\cite{licvpr23amt} and STMFNet~\cite{Danier_2022_CVPR}, using their publicly available implementations. Additionally, we include LDMVFI~\cite{danier2023ldmvfi} and refer to the data reported in paper~\cite{danier2023ldmvfi}. Our PerVFI uses RAFT~\cite{raft} as motion estimator, and we sample the latent code $z$ using temperature $\tau=0.3$. 

\paragraph{Datasets.} To evaluate the performance of the models, we employ commonly used VFI benchmarks: Vimeo-90K~\cite{toflow}, DAVIS (2017)~\cite{davis} and Xiph~\cite{softmaxsplat2020} datasets. 
We opted for video datasets to evaluate perceptual metrics like VFIPS~\cite{vfips} and FloLPIPS~\cite{flolpips} (two bespoke VFI metrics). 
Evaluations for the DAVIS dataset are conducted at both 480P ($640\times 480$) and 1080P ($1920\times 1080$) resolutions. For the Xiph dataset, 8 video sequences, each containing 101 4K frames, are used. Following the approach in~\cite{softmaxsplat2020}, we resize the 4K frames to 2K ($2048\times 1080$) or extract a 2K center crop. All video sequences interpolate even frames based on the corresponding odd frames. Results of comparisons on Middlebury~\cite{middlebury} and UCF101~\cite{ucf101} datasets are included in the appendix. 

\paragraph{Metrics.}
We mainly employ the following metrics for performance evaluation: LPIPS~\cite{lpips} and DISTS~\cite{dists} (image quality metric); VFIPS~\cite{vfips} and FloLPIPS~\cite{flolpips} (video quality metric). These metrics have demonstrated a stronger correlation with human judgments of frame interpolation quality. 
For completeness, we also present the performance of traditional metrics PSNR and SSIM~\cite{ssim}. However, it is important to note that they are not the primary focus of this paper. Higher values indicate better performance for PSNR, SSIM, and VFIPS, while lower values indicate better results for LPIPS and FloLPIPS. Results of comparisons using DISTS~\cite{dists} are included in the appendix.

\begin{table*}[tbp]
  \centering
  \caption{Performance comparison of VFI algorithms on DAVIS-2017~\cite{davis}. The scores for LDMVFI~\cite{danier2023ldmvfi} are taken from their paper and indicated with the $\dagger$ symbol. `OOM' means out of memory. The best values are highlighted in \red{red} and the second-best values are in \blue{blue}.}
  \label{tab:davis}
  \vspace{-5pt}
  \begin{adjustbox}{max width=\textwidth}
  \begin{tabular}{lcccccccccc} 
    \toprule
    {} & \multicolumn{5}{c}{{DAVIS (480P)}} & \multicolumn{5}{c}{{DAVIS (1080P)}} \\
    \cmidrule(lr){2-6} \cmidrule(lr){7-11}
    & {PSNR}$\uparrow$ & SSIM $\uparrow$ & {LPIPS}$\downarrow$ & {FloLPIPS$\downarrow$} & {VFIPS$\uparrow$} & {PSNR$\uparrow$} & SSIM$\uparrow$ & {LPIPS}$\downarrow$ & {FloLPIPS}$\downarrow$ & {VFIPS}$\uparrow$ \\
    \midrule
    EDSC~\cite{cheng2021multiple} & 26.52 & 0.784 & 0.132 & 0.093 & 72.62 & 24.54 & 0.768 & 0.205 & 0.138 & 51.05 \\
    RIFE~\cite{huang2020rife} & 26.97 & 0.807 & \blue{0.085} & \blue{0.063} & \blue{80.19} & 25.89 & 0.803 & \blue{0.134} & \blue{0.097} & \blue{62.56}\\
    STMFNet~\cite{Danier_2022_CVPR} & \blue{28.55} & \blue{0.850} & 0.121 & 0.086 & 77.38 & \blue{27.43} & \blue{0.844} & 0.178 & 0.119 & 60.25\\
    LDMVFI~\cite{danier2023ldmvfi} & 25.54 $^\dagger$ & -   & 0.107 $^\dagger$ & 0.153 $^\dagger$ & 75.78 $^\dagger$ & - & - & - & - & -\\
    VFIFormer~\cite{vfiformer} & 27.33 & 0.814 &  0.124  &  0.090 &  77.32 & OOM & OOM & OOM & OOM & OOM\\
    EMA-VFI~\cite{zhang2023extracting} & \red{28.83} & \red{0.856} &  0.127 &  0.085 &  78.84 & \red{27.61} & \red{0.846} &  0.203 &  0.131 &  60.87   \\
    AMT~\cite{licvpr23amt} & 27.42 & 0.818 &  0.101 &  0.073 &  80.57 & 25.72 & 0.806 &  0.177 &  0.122 &  60.39 \\
    PerVFI (ours) & 26.83 & 0.804 &  \red{0.077} &  \red{0.058} &  \red{87.51} & 26.23 & 0.808 &  \red{0.114} &  \red{0.087} &  \red{72.52}  \\
    \bottomrule
  \end{tabular}
  \end{adjustbox}
\end{table*}

\begin{table*}[tbp]
  \centering
  \caption{Performance comparison of VFI algorithms on Xiph4K~\cite{softmaxsplat2020} and Vimeo-90K~\cite{vimeo}. The best values are highlighted in \red{red}, while the second-best values are in \blue{blue}. `OOM' means out of memory.}
  \label{tab:others}
  \vspace{-5pt}
  \begin{adjustbox}{max width=\textwidth}
  \begin{tabular}{lccccccccc} 
    \toprule
    {} & \multicolumn{3}{c}{{Xiph - 2K}} & \multicolumn{3}{c}{{Xiph - ``4K''}} & \multicolumn{3}{c}{{Vimeo-90K}} \\
    \cmidrule(lr){2-4} \cmidrule(lr){5-7} \cmidrule(lr){8-10}
    & {LPIPS$\downarrow$} & {FloLPIPS$\downarrow$} & {VFIPS$\uparrow$} & {LPIPS$\downarrow$} & {FloLPIPS$\downarrow$} & {VFIPS$\uparrow$} & {PSNR$\uparrow$} & {SSIM$\uparrow$} & {LPIPS$\downarrow$} \\
    \midrule
    EDSC~\cite{cheng2021multiple} & 0.085 & 0.072 & 64.73 & 0.177 & 0.120 & 51.24 & 34.86 & 0.961 & 0.027 \\
    RIFE~\cite{huang2020rife} & \blue{0.041} & \blue{0.050} & 65.26 &  \blue{0.099} & \blue{0.067} & \blue{54.31} & 34.16 & 0.955 & \blue{0.020}\\
    STMFNet~\cite{Danier_2022_CVPR} & 0.110 & 0.063 & 65.19 & 0.245 & 0.128 & 53.33 & - & - & -\\
    VFIFormer~\cite{vfiformer} & OOM & OOM & OOM & OOM & OOM & OOM & \red{36.38} & \red{0.971} & 0.021 \\
    EMA-VFI~\cite{zhang2023extracting} & 0.110 & 0.081 & 65.12 & 0.241 & 0.114 & 53.57 & \blue{36.34} & 0.967 & 0.026\\
    AMT~\cite{licvpr23amt} & 0.089 & 0.055 &  65.60 & 0.199 & 0.114 & 53.22 & 35.79 & \blue{0.968} &  0.021\\
    PerVFI (ours) & \red{0.038} & \red{0.032} & \red{68.67} & \red{0.086} & \red{0.062} & \red{57.47} & 33.89 & 0.953 &  \red{0.018}\\
    \bottomrule
  \end{tabular}
  \end{adjustbox}
  \vspace{-5pt}
\end{table*}

\Subsection{Quantitative Evaluation}
\label{subsec:quantitative}
We present a comparative analysis of PerVFI against state-of-the-art methods in Table~\ref{tab:davis} and~\ref{tab:others}. While STMFNet~\cite{Danier_2022_CVPR} achieves the highest PSNR and SSIM values, its performance falls short in perceptual quality according to the other three metrics. 
Notably, the utilization of symmetric blending in all methods, except for PerVFI, results in noticeable ghosting and blur artifacts. This is observed even in methods with advanced synthesis modules such as the GAN-based STMFNet~\cite{Danier_2022_CVPR} and the diffusion-based LDMVFI~\cite{danier2023ldmvfi}.
By leveraging the ASB module, our PerVFI surpasses other methods significantly in terms of perceptual quality. 
Moreover, when comparing high resolution videos, our PerVFI, trained exclusively on the Vimeo90K dataset, maintains superior visual quality even when compared to STMFNet~\cite{Danier_2022_CVPR} and LDMVFI~\cite{danier2023ldmvfi}, which are trained on high-resolution datasets. 
This further showcases the generalization capability of our proposed method.

\begin{figure}
    \centering
    \includegraphics[width=\linewidth]{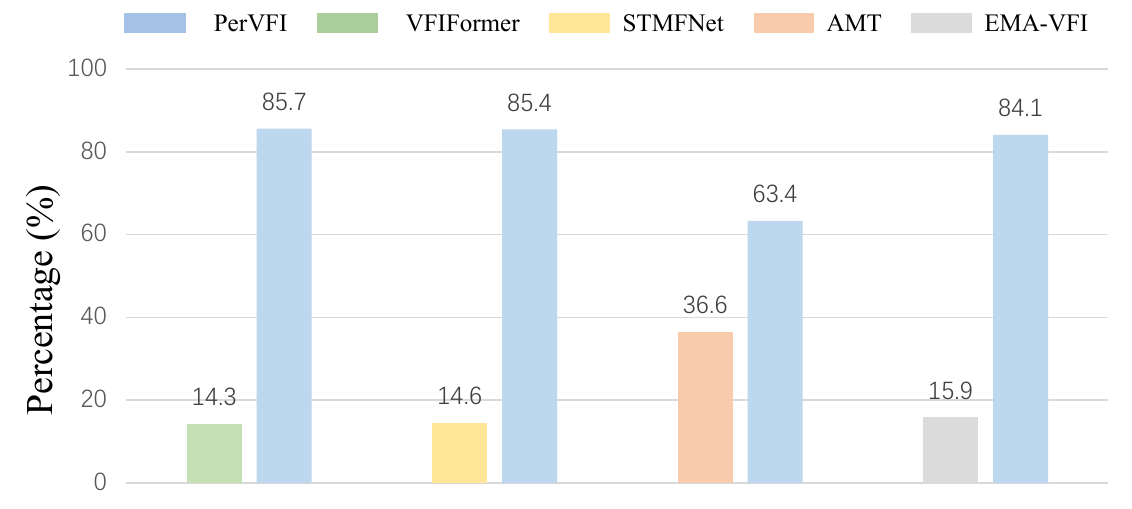}
    \caption{User study results.}
    \label{fig:user}
    \vspace{-10pt}
\end{figure}

\begin{figure*}[tbp]
    \centering
    \includegraphics[width=\linewidth]{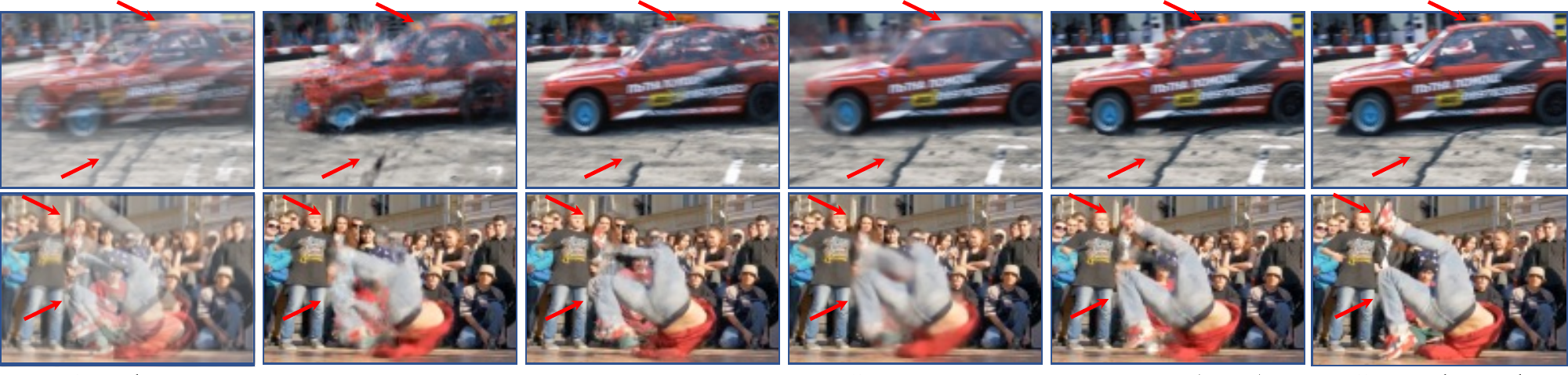}
    \begin{picture}(0,0)
        \put(-40,5){\makebox(0,0)[c]{\small RIFE~\cite{huang2020rife}}}
        \put(-122,5){\makebox(0,0)[c]{\small EMA-VFI~\cite{zhang2023extracting}}}
        \put(-205,5){\makebox(0,0)[c]{\small Overlays}}
        \put(43,5){\makebox(0,0)[c]{\small STMFNet~\cite{Danier_2022_CVPR}}}
        \put(125,5){\makebox(0,0)[c]{\small PerVFI (ours)}}
        \put(210,5){\makebox(0,0)[c]{\small Ground Truth}}
    \end{picture}
    \vspace{-5pt}
    \caption{Perceptual quality comparison between different methods. Our approach produces a high-quality result in spite of the fast-moving objects that is subject to large motion. Red arrows emphasize areas where PerVFI excels in visual quality compared to other methods.}
    \vspace{-15pt}
    \label{fig:visual}
\end{figure*}

\Subsection{Qualitative Evaluation}
\label{subsec:qualitative}
We also compare the visual quality of different methods in DAVIS dataset, as illustrated in Figure~\ref{fig:visual}. 
In regions with significant pixel displacement, it is evident that other methods produce outputs with noticeable ghosting artifacts or blurriness, resulting in a significant degradation of visual quality. In contrast, PerVFI consistently generates outputs with sharp edges and intact content, leading to visually pleasing results. It is important to note that while the PerVFI results may not exhibit perfect pixel-wise alignment with the ground-truth image, the overall visual quality remains consistent with the reference images. 

Additionally, to facilitate a more comprehensive examination of the visual quality, we conduct a user study involved 33 participants comparing 98 videos interpolated with 5 methods. We conduct A/B test on perceptual quality, where the ratio values indicate the percentages of participants preferring the corresponding model. Statistical results are presented in Figure~\ref{fig:user}, demonstrating our method consistently outperforms others.

\begin{figure}[tbp]
    \centering
    \includegraphics[width=\linewidth,height=0.8\linewidth]{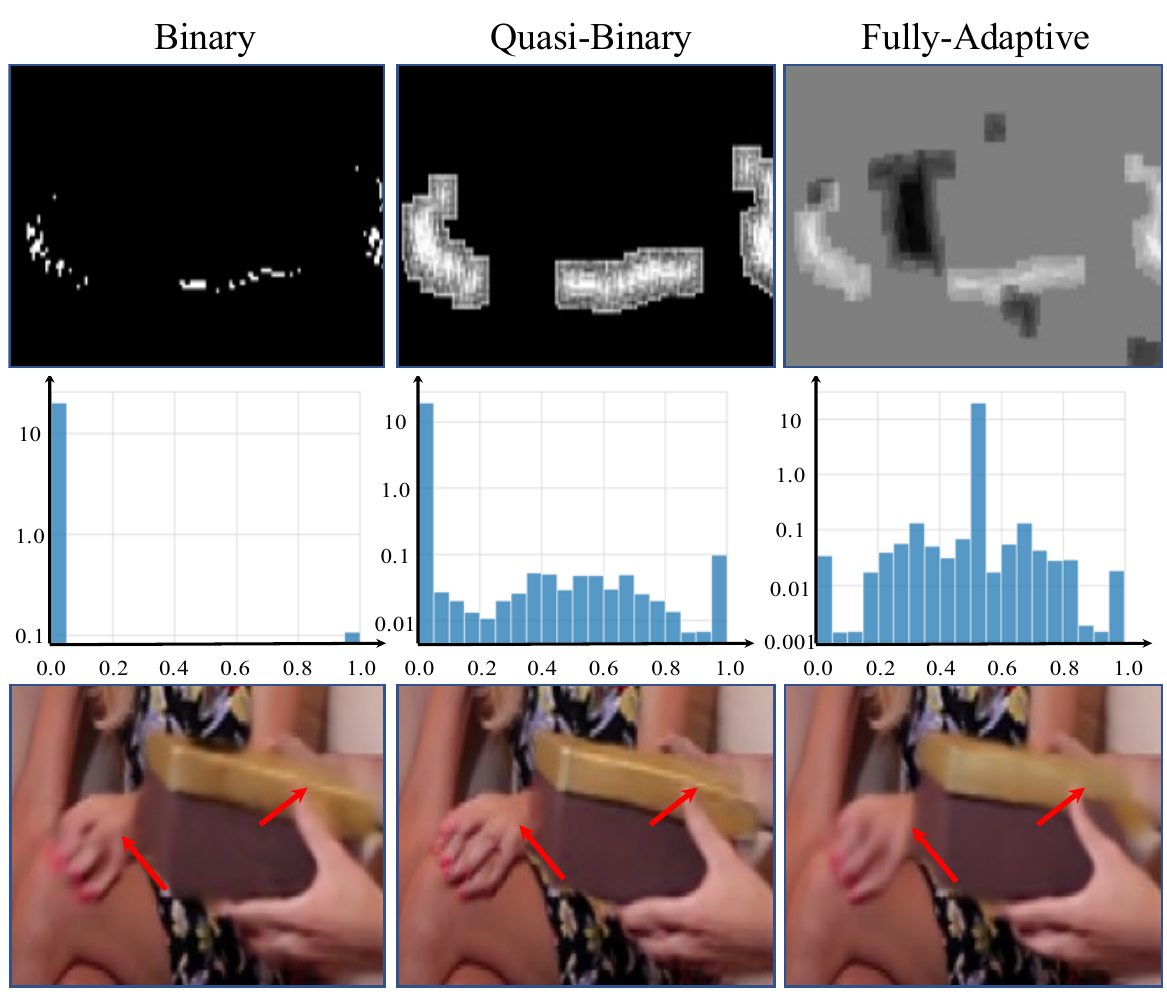}
    \caption{The first row visualize different masks, and the second row exhibits histograms for each. The fully-adaptive mask tends to center around 0.5, signifying an equal contribution from both sides features. The quasi-binary mask maintains sparsity while being partially adaptive, providing an effective blending mechanism. The third row presents results using these masks. Red arrows emphasize areas where the quasi-binary mask excels in visual quality}
    \label{fig:mask}
    \vspace{-8pt}
\end{figure}

\Subsection{Ablation Experiments}
\label{subsec:ablation}






  \begin{table}[tbp]
    \centering
    \caption{Ablation Experiment Results. We show the PSNR and VFIPS on DAVIS (480P) dataset.}
    \label{tab:ablation_results}
    \vspace{-5pt}
    \begin{adjustbox}{max width=\textwidth}
    \begin{tabular}{cccccc}
      \toprule
      Mask & Noise & Prior & Loss & PSNR & VFIPS \\
      \midrule
      Binary & \ding{55} & \ding{51} & Bi-D. & 26.52 & 81.01 \\
      Quasi-B. & \ding{55} & \ding{51} & Bi-D. & 26.71 & 81.24 \\
      Quasi-B. & \ding{51} & \ding{55} & Bi-D. & 27.10 & 82.27 \\
      Quasi-B. & \ding{51} & \ding{55} & L1 & 26.81 & 80.34 \\
      Quasi-B. & \ding{51} & \ding{51} & L1 & 27.15 & 80.50 \\
      Quasi-B. & \ding{51} & \ding{51} & NLL & 26.98 & 81.88 \\
      Binary & \ding{51} & \ding{51} & Bi-D. & 26.69 & 81.25 \\
      Adaptive & \ding{51} & \ding{51} & Bi-D. & \textbf{27.61} & 78.20 \\
      \midrule
      Quasi-B. & \ding{51} & \ding{51} & Bi-D. & 27.16 & \textbf{83.30} \\
      \bottomrule
    \end{tabular}
    \end{adjustbox}
\end{table}%

\begin{figure}[tbp]
    \centering
    \includegraphics[width=0.8\linewidth]{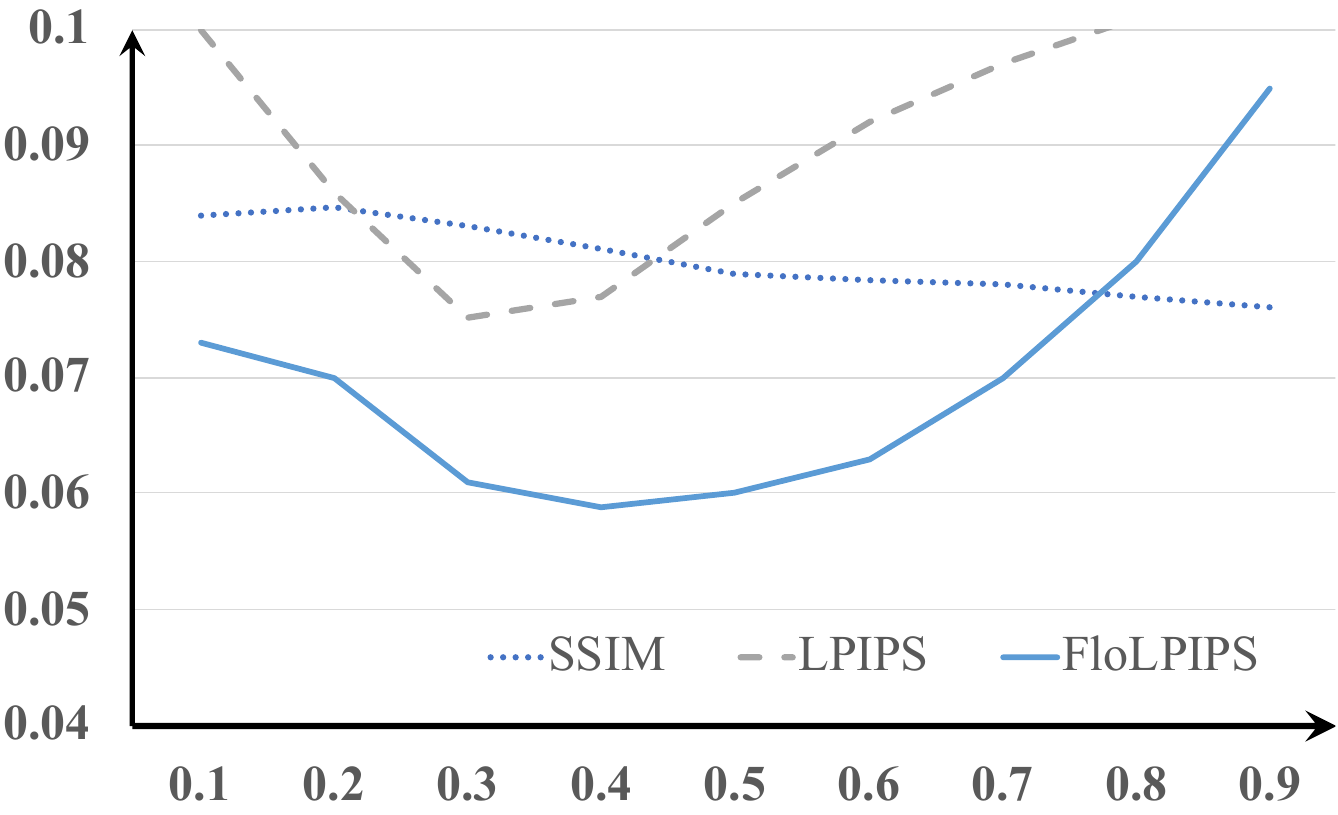}
    \caption{Different $\tau$ during inference. The SSIM is divided with 10 to enhance visibility.}
    \label{fig:curve}
    \vspace{-10pt}
\end{figure}

\paragraph{Symmetric vs. Asymmetric blending.}
To emphasize the importance of asymmetric synergistic blending, we have designed a symmetric blending module for a fair comparison. In this module, we introduce a learnable bias after each convolution layer in the Adaptive Blending Module (ADM), allowing the output mask to be fully adaptive without any constraints.
As depicted in the first row of Figure~\ref{fig:mask}-(a), the fully adaptive mask tends to merge features from both sides equally, resulting in a blurry output, as shown in the second row. Furthermore, in Figure~\ref{fig:mask}-(b), we provide a histogram illustrating the distribution of values for different types of masks. The quasi-binary mask (denoted as Quasi-B.) maintains overall sparsity while being partially adaptive.
In Table~\ref{tab:ablation_results}, we use ``Adaptive'' to indicate the fully adaptive mask without any constraints, and ``Quasi-B.'' to represent the quasi-binary mask. As observed, symmetric blending with a fully adaptive weighting mask achieves higher PSNR values but lower VFIPS scores compared to the asymmetric blending with the quasi-binary mask. This phenomenon demonstrates the significance of imposing strict constraints during the blending process to enhance the perceptual quality of the output.

\paragraph{Quasi-binary mask.}
In the ADM, we introduce a random uniform noise term $n$ during training. This is done because multiplying with a sparse mask could potentially lead to gradient vanishing issues. To showcase the importance of this operation, we conduct experiments where we train the model without the adding the noise. Specifically, we evaluate the model using either the quasi-binary mask or a binary mask without dilation.
As presented in the first and second rows of Table~\ref{tab:ablation_results}, the performance of the model experiences a noticeable degradation when trained without the inclusion of random noise. This emphasizes the necessity of incorporating the random noise term during training. We will provide further visual comparisons in the appendix to complement these findings.

\paragraph{PAM module.}
In the PAM, we utilize the optical flow $\mathbf{F}_{1\rightarrow 0}$ as prior information for alignment. We found it necessary to incorporate this prior in order to effectively handle occlusion compensation. As demonstrated in Table~\ref{tab:ablation_results}, removing prior information from the PAM leads to failures in occlusion compensation in the resulting frames.
Additional visual comparisons are shown in the appendix, highlighting the impact of the prior information on the quality of the interpolated frames.

\paragraph{Different loss functions.}
In Table~\ref{tab:ablation_results}, we also compare different loss functions in the training stage. Our PerVFI employs the bi-directional loss (denoted as Bi-D) as described in Equation~(\ref{eq:loss}). For comparison, we also train the network only using the negative log-likelihood (NLL) loss or the L1 loss.
We have observed that training with only the NLL loss introduces certain noise in the output frames, particularly in misaligned regions. However, by introducing the perceptual loss, we are able to suppress this noise and generate better results. As shown in table~\ref{tab:ablation_results}, the L1 loss yields higher PSNR but lower VFIPS. 
To further highlight the superiority of our loss function, we provide additional visual samples in the appendix, demonstrating the effectiveness of our approach in producing visually superior results.

\paragraph{Different variances for latent codes.}
In Figure~\ref{fig:curve}, 
we present results by sampling the latent codes with different variances $\tau$ during the inference stage. 
We can observe that the three metrics (SSIM, LPIPS and FloLPIPS) exhibit different optimal ranges. It is worth noting that the latent code can be flexibly customized to strike a balance between different metrics based on specific preferences.





\section{Conclusion}
We introduce PerVFI, a novel approach for video frame interpolation that decisively tackles issues of blur and ghosting artifacts, thereby significantly elevating perceptual quality. Our design incorporates an asymmetric blending module that strategically leverages features from two reference frames: one for primary content and the other for occlusion information. The model employs a normalizing flow-based generator with negative log-likelihood loss to capture the latent conditional distribution. The experiments validate its superiority in artifact reduction and high-quality generation.

\section{Acknowledgment}
The work was supported by the National Natural Science Foundation of China (62301310, U23A20391), the Shanghai Pujiang Program (22PJ1406800), and the GuangDong Basic and Applied Basic Research Foundation (2023A1515010644, 2021B151520011).


\clearpage\newpage
\medskip
{
    \small
    \bibliographystyle{ieeenat_fullname}
    \bibliography{main}
}

\end{document}